\documentclass{exam}

\usepackage[utf8]{inputenc} 
\usepackage[T1]{fontenc}    
\usepackage{hyperref}       
\usepackage{url}            
\usepackage{booktabs}       
\usepackage{amsfonts}       
\usepackage{nicefrac}       
\usepackage{microtype}      
\usepackage{float}
\usepackage{amsmath}
\usepackage{tikz}
\usepackage{graphicx}
\usepackage{adjustbox}
\usepackage{indentfirst}
\usepackage{multicol}
\usepackage{titling}
\usepackage[margin=0.75in]{geometry}
\usepackage{sectsty}
\usepackage[nodisplayskipstretch]{setspace}
\sectionfont{\fontsize{12}{12}\selectfont}
\usepackage{endnotes}
\usepackage{stfloats}

\usepackage{xparse}
\NewDocumentCommand{\codeword}{v}{%
    \texttt{{#1}}%
}

\let\footnote=\endnote

\title{Achieving Competitive Play Through Bottom-Up Approach in Semantic Segmentation}
\author{
\small Eric Pryzant \\ \small \href{mailto:eric.pryzant@utexas.edu}{eric.pryzant@utexas.edu} 
\and \small Qicong (Alvin) Deng \\ \small \href{mailto:alvin.q.deng@utexas.edu}{alvin.q.deng@utexas.edu}
\and \small Bill Mei \\ \small \href{mailto:billmei@utexas.edu}{billmei@utexas.edu} 
\and \small Evan Shrestha \\ \small \href{mailto:evanshrestha@utexas.edu}{evanshrestha@utexas.edu} 
}
\date{\vspace{-5ex}} 

\begin{document}
\maketitle
\begin{figure*}[!b]
    \centering
    \includegraphics[scale=0.16]{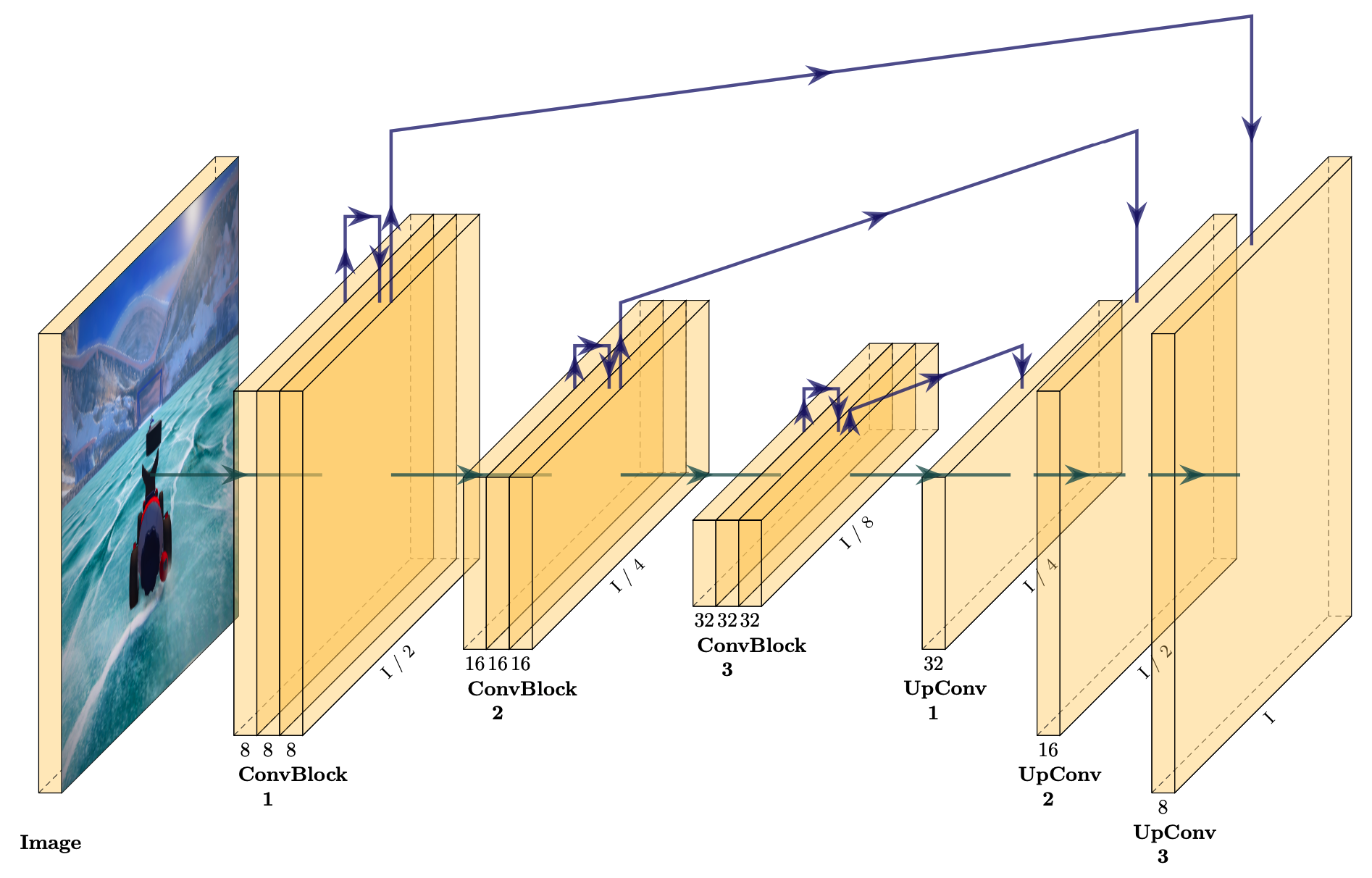}
    \caption{PuckNet architecture. The green arrow represents the direct connection between layers. The blue arrow denotes an residual/skip connection between layers. The length of each layer represents the number of channels. The architecture goes from 8, 16, 32 channels (from the convolutions) down to 32, 16, 8 channels (from the up-convolutions).}
\end{figure*}
\begin{multicols}{2}

\section{Abstract}
With the renaissance of neural networks, object detection has slowly shifted from a bottom-up recognition problem to a top-down approach. Best in class algorithms enumerate a near-complete list of objects and classify each into  object/not object. In this paper, we show that strong performance can still be achieved using a bottom-up approach for vision-based object recognition tasks and achieve competitive video game play. We propose PuckNet, which is used to detect four extreme points (top, left, bottom, and right-most points) and one center point of objects using a fully convolutional neural network. Object detection is then a purely keypoint-based appearance estimation problem, without implicit feature learning or region classification. The method proposed herein performs on-par with the best in class region-based detection methods, with a bounding box AP of $36.4\%$ on COCO test-dev. In addition, the extreme points estimated directly resolve into a rectangular object mask, with a COCO Mask AP of $17.6\%$, outperforming the Mask AP of vanilla bounding boxes. Guided segmentation of extreme points further improves this to $32.1\%$ Mask AP. We applied the PuckNet vision system to the SuperTuxKart video game to test it's capacity to achieve competitive play in dynamic and co-operative multiplayer environments.

\section{Vision System}

The primary task is to use binary classification to detect if a puck is in the image, and the second task is to use regression to predict the coordinate of the puck.

\subsection{Model Architecture}
We designed a Fully Convolutional Network (FCN) similar to an U-Net architecture \cite{ronneberger2015u} with residual connections from ResNet \cite{he2016deep}. The architecture first uses three down-convolutional blocks where each block has three 3x3 convolutions followed by batch normalization and ReLU.  Next, the architecture uses three up-convolutions to rescale the feature map back to the original dimension. We used residual connections between the down-blocks and the up-blocks to avoid vanishing gradients. Borrowing ideas from U-Net, the architecture uses skip connections between each pair of down-convolution and up-convolution to leverage the low-level feature map to assist high-level features in terms of restoring the image details due to the coordinate prediction. The architecture at the end uses the last up-convolution to generate the puck classification by applying a global average pooling followed by a linear unit as well as the coordinate regression by using a 1x1 convolution along with a Softmax operation followed by a scaling operation.

\subsection{Task Input}
Within the \codeword{act} function in \codeword{agent/player.py}, we leveraged the in-game image $I \in R^{3 \times W \times H}$, where W and H are the width and height of the image with RGB channels, as the input data for both classification and regression tasks.

\subsection{Task Label}
Through \codeword{pystk}'s API\footnote{\url{https://pystk.readthedocs.io/en/latest/}}, we extracted the segmentation mask of the puck of each image and its in-game coordinate as the labels for the puck classification and coordinate regression.

To generate the binary label of the puck classification, we used the following logic.

\setstretch{0.1}
\begin{equation}
L_{I} =
\begin{cases}
  1, & \text{if}\ PuckPixelCount(M_{I}) > 20 \\
  0, & \text{otherwise}
\end{cases}
\end{equation}
\setstretch{1.0} $PuckPixelCount(M_{I})$ is a function that counts the number of puck pixels in the segmentation mask $M_{I}$ of image $I$. If the number of puck pixels is greater than 20, then the label $L_{I}$ of the image $I$ is positive and vice versa. One caveat is that the pucks that are far-away from the camera are hard to recognize. Therefore, we chose the threshold to be at 20 pixels to ensure the model learns to recognize reasonably-sized pucks. This task is similar to the classification task in standard vision systems.

To generate the label of the coordinate regression, we used a utility function called \codeword{_to_image} to convert the in-game $x, y, z$ coordinates to the on-screen $x, y$ coordinates where $0 \le x \le 400$ and $0 \le y \le 300$. This task is very similar to the aim-point regression task in segmented regression networks.

\section{Agent Design}
\small
Our general strategy is to have one agent focus on scoring the puck into the goal (the ``Chaser''), leaving the other agent to focus on tracking the puck location as accurately as possible (the ``Spotter'') and communicating this information to the first agent. This strategy is built around the idea that one agent with perfect information would perform better than two with imperfect information.

We play-tested the game using the online multiplayer option against ourselves for several rounds to understand the game mechanics and come up with potential strategies. We noticed that some characters were easier to drive than others; to pick the best character, we set up a script to play 2,210 games against the AI (17 characters times 130 rounds each). The results are summarized in Table~\ref{kart_character:table}, and we picked Xue as the Chaser.

\subsection{Spotter}
The Spotter tries to keep the puck in sight at all times. The Spotter starts the match by rushing to the middle as fast as possible to get a lock on the puck. Once the puck is found, the agent switches to a `watching' state and waits until the puck moves. As the puck moves, the Spotter tries to maintain a fixed distance away from it. We found that turning backwards was much quicker than going forwards, so we made this the primary method for tracking the puck when it was close to the agent. If the puck moved further away, the agent would instead enter a ``chase'' state that would help it keep the puck within a set distance. We  implemented several other states with custom logic to help the agent deal with adverse situations, such as getting blown up by an enemy bomb, getting knocked off course by an enemy agent, and also what to do when the puck is out of sight.

We play-tested every kart character and picked the best Spotter based on the kart's acceleration, turning radius, and viewing angle. We picked Wilber as our Spotter that best fulfilled the above criteria.

\subsection{Chaser}

\begin{figure}[H]
    \centering

    \tikzset{every picture/.style={line width=0.75pt}} 
    \begin{tikzpicture}[x=0.75pt,y=0.75pt,yscale=-1,xscale=1]
    
    \draw  [draw opacity=0][fill={rgb, 255:red, 144; green, 19; blue, 254 }  ,fill opacity=1 ] (175.75,90.42) -- (183.79,93.46) -- (189.46,87.31) -- (186.59,95.17) -- (192.97,100.93) -- (184.93,97.88) -- (179.26,104.03) -- (182.14,96.17) -- cycle ;
    \draw   (21.27,35.8) .. controls (21.27,23.21) and (31.47,13) .. (44.07,13) -- (241.36,13) .. controls (253.96,13) and (264.16,23.21) .. (264.16,35.8) -- (264.16,104.2) .. controls (264.16,116.79) and (253.96,127) .. (241.36,127) -- (44.07,127) .. controls (31.47,127) and (21.27,116.79) .. (21.27,104.2) -- cycle ;
    \draw   (184.17,91.7) .. controls (184.17,85.8) and (188.96,81.01) .. (194.86,81.01) .. controls (200.76,81.01) and (205.55,85.8) .. (205.55,91.7) .. controls (205.55,97.6) and (200.76,102.39) .. (194.86,102.39) .. controls (188.96,102.39) and (184.17,97.6) .. (184.17,91.7) -- cycle ;
    \draw    (264.16,69.35) -- (187.04,94.64) ;
    \draw [shift={(185.14,95.26)}, rotate = 341.85] [color={rgb, 255:red, 0; green, 0; blue, 0 }  ][line width=0.75]    (10.93,-3.29) .. controls (6.95,-1.4) and (3.31,-0.3) .. (0,0) .. controls (3.31,0.3) and (6.95,1.4) .. (10.93,3.29)   ;
    
    \draw  [fill={rgb, 255:red, 74; green, 144; blue, 226 }  ,fill opacity=1 ] (261.25,69.35) .. controls (261.25,67.74) and (262.55,66.44) .. (264.16,66.44) .. controls (265.77,66.44) and (267.08,67.74) .. (267.08,69.35) .. controls (267.08,70.96) and (265.77,72.27) .. (264.16,72.27) .. controls (262.55,72.27) and (261.25,70.96) .. (261.25,69.35) -- cycle ;
    \draw  [fill={rgb, 255:red, 208; green, 2; blue, 27 }  ,fill opacity=1 ] (19,69.35) .. controls (19,67.74) and (20.3,66.44) .. (21.91,66.44) .. controls (23.52,66.44) and (24.83,67.74) .. (24.83,69.35) .. controls (24.83,70.96) and (23.52,72.27) .. (21.91,72.27) .. controls (20.3,72.27) and (19,70.96) .. (19,69.35) -- cycle ;
    
    \draw (58,66) node  [color={rgb, 255:red, 208; green, 2; blue, 27 }  ,opacity=1 ] [align=left] {Our Goal};
    \draw (208,63.2) node  [color={rgb, 255:red, 74; green, 144; blue, 226 }  ,opacity=1 ] [align=right] {Opponent's Goal};
    \draw (225,103) node   [align=left] {Puck};
    \draw (125,96) node  [color={rgb, 255:red, 144; green, 19; blue, 254 }  ,opacity=1 ] [align=right] {Target Location};
    \end{tikzpicture}
    \caption{Targeting Logic for the Spotter Agent}
    \label{fig:target_star}
\end{figure}
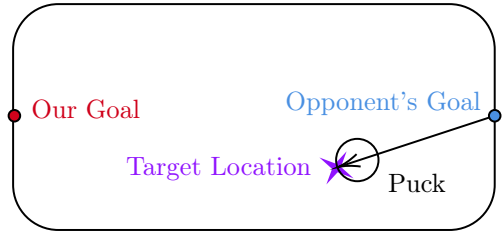

To position Chaser in the right place to score goals, we project a ray from the opponent's goal to 20 length units behind the puck as the target location that Chaser tries to drive to. Figure \ref{fig:target_star} visualizes this target location as a purple star.

Once the Chaser drives to the target location, scoring a goal becomes a simple matter of successively updating the target location for the agent to push the puck as it glides into the opponent's net.


\subsection{Alternative Strategies Considered}
We considered using reinforcement learning for this task but ruled it out because the reward for scoring is too infrequent and it would require hundreds of hours of GPU computation to train. We considered imitation learning based on the game's built-in AI, but in our play-testing, the AI frequently own-goaled, so we did not think its strategy was worth copying. We considered a targeting strategy that would predict the puck's future location based on its current velocity but decided that this was too difficult because our prediction is noisy. If the prediction is off by even a little bit, the velocity vector can point in wildly different directions. We considered training an additional neural net to identify opponent karts to launch items that the agent picks up during play. However, we found that scoring against the AI was already difficult, so we decided not to distract our agent from the primary goal. We also noticed that our aim was rarely on target, so the benefit was minimal. We considered giving the Spotter additional duties as a defender so it would park itself between the puck and our goal once it had a lock on the puck. This secondary objective, however, reduced its ability to react quickly when the puck moved and significantly hindered the agent's ability to track the puck, so we decided to keep it focused on its primary objective. Finally, we also looked into creative turning strategies. We discovered it was much faster to turn around by going behind the puck and then calling a rescue which turns the player to face the blue goal. We did not implement this strategy in our agent because it only works if the agents are on the red team.

\section{Data Collection}
We designed agents \codeword{zamboni}, \codeword{spotter_gather}, and \codeword{spotter_trainer} for data collection. We ran the agents with ground truth data from \codeword{pystk} under combinations of settings such as different starting team and karts to improve the diversity of the dataset and the robustness of the model. We collected a total of 280,000 examples and split them into 70\% training data for model learning, 15\% validation data for hyper-parameter tuning, and 15\% test data for generalization measures.

The \codeword{zamboni} agent was placed evenly in a grid around the map, facing all four cardinal directions to collect image data from all locations. At each location, the puck and three other AI agents were randomly placed on the screen using the \codeword{to_world} function, which was adapted from previous work on segmented regression.

The \codeword{spotter_gather} and \codeword{spotter_trainer} agents were used to collect natural images, such as when the puck is shadowed along the wall of the rink. Both agents were extensions of the \codeword{spotter} agent. These agents reset after a certain number of frames passed. After a reset, the puck and karts were placed randomly, and the agent followed the puck using the logic from the \codeword{spotter} agent. With this method, we were able to collect data with natural variation, such as non-cardinal rotations, projectiles, karts, and shadows along the walls.

\section{Experiment}
During training we optimize the following, multi-part loss function:
\setstretch{0.1}
\begin{equation}
    \ell_{class}(l, c) = \frac{1}{N} \sum_{i=1}^{N}[c_{i} \cdot \log \sigma(l_{i}) + (1 - c_{i}) \cdot \log(1 - \sigma(l_{i}))]
\end{equation}
\setstretch{1.0}where $N$ denotes the number of samples, $c_{i}$ denotes the puck label, and $l_{i}$ denotes the predicted raw logits from the model. The first part of the loss function is the standard binary cross-entropy loss.
\setstretch{0.1}
\begin{equation}
    \ell_{reg}(s, y) = \frac{1}{N} \sum_{i} z_{i}
\end{equation} where $z_{i}$ is given by

\begin{equation}
    z_{i} =
        \begin{cases}
        0.5 (s_i - y_i)^2, & \text{if } |s_i - y_i| < 1 \\
        |s_i - y_i| - 0.5, & \text{otherwise}
        \end{cases}
\end{equation} \setstretch{1.0} where $y_{i}$ denotes the ground truth of puck coordinates, and $s_{i}$ denotes the predicted coordinates from the model. The second part of the loss function is known as the Smooth L1 loss \cite{girshick2015fast} or the Huber loss\footnote{\url{https://en.wikipedia.org/wiki/Huber_loss}}. This loss is known to be less sensitive to outliers than typical loss functions such as MSE loss and in some situations, prevents exploding gradients.
\setstretch{0.1}
\begin{equation}
    \ell(l, c, s, y) = \ell_{class}(l, c) + \ell_{reg}(s, y)
\end{equation} 
\setstretch{1.0} where the overall loss function we used is a combination of Smooth L1 loss for the coordinate regression and binary cross-entropy loss for the puck classification.

\subsubsection{Regression Masking}
In situations where the puck is not in the field of view of the camera, we applied a masking trick where we multiply the puck classification label with the Smooth L1 loss. The idea is that if the classification label is negative (the label is 0), then the Smooth L1 loss cancels out, so the model is not learning on any faulty coordinate labels when the puck is not in view.

\subsection{Training}
We implemented the model in PyTorch \cite{NEURIPS2019_9015} and trained each model using the Adam \cite{kingma2014adam} optimizer with a learning rate of 1e-3 and a batch size of 128 samples due to GPU memory constraints. We also applied an L2 regularization to the model weights and an image augmentation strategy to improve the robustness of the model. The image augmentation pipeline consists of randomly flipping the images horizontally\footnote{\url{https://pytorch.org/docs/stable/torchvision/transforms.html\#torchvision.transforms.RandomHorizontalFlip}} and randomly changing the brightness, contrast, saturation, and the hue of images.\footnote{\url{https://pytorch.org/docs/stable/torchvision/transforms.html\#torchvision.transforms.ColorJitter}} We trained each model up to 100 epochs with an early stopping with a tolerance of 5 epochs on the validation loss to prevent over-fitting.

\subsection{Evaluation Method}
We used ROC-AUC\footnote{\url{https://scikit-learn.org/stable/modules/model_evaluation.html\#roc-metrics}} and PR-AUC\footnote{\url{https://scikit-learn.org/stable/modules/model_evaluation.html\#precision-recall-f-measure-metrics}} for classification tasks and mean absolute error (MAE)\footnote{\url{https://scikit-learn.org/stable/modules/model_evaluation.html\#mean-absolute-error}} for regression tasks as metrics to evaluate our model. For the classification metrics, an ROC curve demonstrates the performance of a binary classifier as its decision threshold is varied, and it is plotting True Positive Rate (TPR) against False Positive Rate (FPR). TPR is defined as $TPR = \frac{TP}{TP + FN}$ and FPR is defined as $FPR = \frac{FP}{FP + TN}$. A PR curve is plotting Precision against Recall, and it is much better in illustrating differences in binary classifiers when there are data imbalances. Precision is defined as $Precision = \frac{TP}{TP + FP}$ and Recall is defined as $Recall = \frac{TP}{TP + FN}$, which is similar to TPR. For regression metrics, MAE is defined as \setstretch{0.1}$$MAE(y, \hat{y}) = \frac{1}{N} \sum_{i=1}^{N} |y_{i} - \hat{y}_{i}|$$ \setstretch{1.0} where $N$ is the number of samples, $\hat{y}_{i}$ is the predicted value, and $y_{i}$ is the ground truth value of the $i$-th sample.

\subsection{Result}
We did a hyper-parameter search with only the L2 regularization strength due to time constraints as well as the latency constraint from the final project requirement. The scan of the L2 regularization strength goes from 1e-3 to 1e-6. The results are summarized in Table~\ref{model_result:table}. Note that the models have very similar performance in terms of ROC-AUC and PR-AUC. To pick the best performing model, we played 40 games against the AI (4 models times 10 rounds each). The results are summarized in Table \ref{model_mock_tournament:table}, and we picked the model with a regularization of 1e-6 as the best performing model.

\end{multicols}

\theendnotes

\bibliographystyle{unsrt}  
\bibliography{references} 

\section{Appendix}

\begin{table}[H]
\centering
\caption{Performance of each kart character with perfect information}
\begin{tabular}{|l|l|l|l|l|l|l|}
\hline

Character & Wins & Ties & Losses & Net & Avg Goals & Avg Net Goals \\ \hline
xue & 55 & 39 & 36 & 19 & 1.10 & 0.15 \\ \hline
amanda & 48 & 42 & 40 & 8 & 1.14 & 0.06 \\ \hline
wilber & 40 & 51 & 40 & 0 & 0.85 & 0.00 \\ \hline
hexley & 37 & 46 & 47 & -10 & 0.48 & -0.08 \\ \hline
gnu & 40 & 38 & 52 & -12 & 0.95 & -0.09 \\ \hline
sara\_the\_racer & 48 & 22 & 60 & -12 & 0.97 & -0.09 \\ \hline
kiki & 43 & 28 & 59 & -16 & 0.69 & -0.12 \\ \hline
suzanne & 35 & 41 & 54 & -19 & 0.64 & -0.15 \\ \hline
tux & 29 & 43 & 58 & -29 & 0.53 & -0.22 \\ \hline
sara\_the\_wizard & 34 & 30 & 66 & -32 & 0.60 & -0.25 \\ \hline
adiumy & 29 & 38 & 63 & -34 & 0.58 & -0.26 \\ \hline
gavroche & 33 & 29 & 68 & -35 & 0.61 & -0.27 \\ \hline
nolok & 35 & 25 & 70 & -35 & 0.70 & -0.27 \\ \hline
pidgin & 33 & 29 & 68 & -35 & 0.58 & -0.27 \\ \hline
emule & 31 & 32 & 67 & -36 & 0.60 & -0.28 \\ \hline
beastie & 26 & 36 & 68 & -42 & 0.60 & -0.32 \\ \hline
puffy & 28 & 27 & 75 & -47 & 0.67 & -0.36 \\ \hline
konqi & 30 & 14 & 86 & -56 & 0.75 & -0.43 \\ \hline

\end{tabular}
\label{kart_character:table}
\end{table}

\begin{table}[H]
\caption{Hyper-parameter search result. The metrics reported are in the following dataset format: Train / Validation / Test}
\begin{adjustbox}{width=\columnwidth,center}
\begin{tabular}{|l|l|l|l|l|}
\hline
Regularization Strength & Loss & ROC-AUC & PR-AUC & MAE \\ \hline
1e-3 & 7.568 / 2475.307 / 2710.843 & 0.995 / 0.995 / 0.995 & 0.997 / 0.996 / 0.997 & 12.798 / 14.149 / 14.006 \\ \hline
1e-4 & 7.333 / 2398.976 / 2642.883 & 0.996 / 0.996 / 0.996 & 0.998 / 0.997 / 0.998 & 12.446 / 13.718 / 13.658 \\ \hline
1e-5 & 7.408 / 2427.712 / 2661.627 & 0.996 / 0.996 / 0.995 & 0.998 / 0.997 / 0.997 & 12.583 / 13.904 / 13.775 \\ \hline
1e-6 & 7.395 / 2425.357 / 2669.386 & 0.996 / 0.996 / 0.996 & 0.998 / 0.997 / 0.998 & 12.590 / 13.873 / 13.800 \\ \hline
\end{tabular}
\label{model_result:table}
\end{adjustbox}
\end{table}

\begin{table}[H]
\centering
\caption{Performance of each vision model}
\begin{tabular}{|l|l|l|l|l|l|}
\hline

Regularization Strength & Wins & Ties & Losses & Goals Scored & Goals Conceded \\ \hline
1e-3 & 0 & 0 & 10 & 0 & 19 \\ \hline
1e-4 & 0 & 2 & 8 & 2 & 19 \\ \hline
1e-5 & 2 & 1 & 7 & 5 & 14 \\ \hline
1e-6 & 2 & 3 & 5 & 5 & 18 \\ \hline
\end{tabular}
\label{model_mock_tournament:table}
\end{table}

\end{document}